\documentclass[preprint,5p,times,twocolumn]{elsarticle}

\usepackage{amssymb}
\usepackage{amsmath}
\usepackage{booktabs}
\usepackage{multirow}
\usepackage{xcolor}
\usepackage{placeins}
\usepackage{dblfloatfix}
\journal{Nuclear Physics B}

\begin{document}

\begin{frontmatter}



\title{DualCast: A Dual-Path Language Model for Bimodal
Financial Time-Series Forecasting}


\author[1,2]{Wentao Zhao$^\ddag$}
\author[3,4]{Hongqiang Wu$^\ddag$}
\author[1]{Shanghang Liu$^\ddag$}

\author[1]{Zhaochen Zan}
\author[3]{Yu Zhang$^\dag$}
\author[1]{Biqing Huang$^\dag$}

\fntext[cor1]{$^\dag$ Corresponding author}
\fntext[equal]{$^\ddag$ Equal contribution.}

\affiliation[1]{
organization={Department of Automation, Tsinghua University},
city={Beijing},
postcode={100084},
country={China}
}
\affiliation[2]{
organization={School of Software, Tsinghua University},
city={Beijing},
postcode={100084},
country={China}
}
\affiliation[3]{
organization={School of Computer Science, Northwestern Polytechnical University},
city={Xi'an},
postcode={710129},
country={China}
}
\affiliation[4]{
organization={Inspur Yunzhou Industrial Internet Co. Ltd},
city={Jinan},
postcode={250101},
country={China}
}

\begin{abstract}
Financial time-series forecasting must capture price dynamics across heterogeneous assets while incorporating news available at prediction time. We introduce DualCast, a dual-path framework that extends a frozen language model with a discrete financial vocabulary. Each log-return patch is represented by a learned summary token and three residual shape tokens, preserving local drift and volatility while allowing shape patterns to be shared across assets. To improve codebook utilization, we develop adaptive frequency-equalizing residual vector quantization, which rebalances overloaded codewords without compromising reconstruction accuracy. The fast path trains only the new financial-token embeddings and output heads on a frozen Qwen3-8B backbone. A toggleable LoRA adapter enables a slow path that conditions on the fast forecast and news available at the forecast origin to produce a revised prediction. The reviser is initialized by supervised fine-tuning and further optimized with a return-space group relative policy optimization objective that rewards improvements over the fast forecast. In zero-shot evaluations covering equities and energy prices at five-minute, daily, and weekly resolutions, the slow path achieves the lowest mean absolute percentage error among the compared methods in 8 of 12 dataset–horizon settings, including every longest-horizon setting. News ablations indicate additional gains in most tested settings, although their magnitude varies across markets. DualCast thus combines a fast numerical forecaster with an optional text-conditioned revision mechanism.

\end{abstract}



\end{frontmatter}

\section{Introduction}

Despite recent progress in applying large language models (LLMs) to time-series forecasting, financial markets remain particularly challenging. First, financial data are highly heterogeneous, exhibit low signal-to-noise ratios, and evolve under rapidly changing volatility regimes, creating a substantial gap from conventional forecasting benchmarks and making stable pattern learning difficult. Second, financial forecasting requires not only modeling historical price dynamics but also understanding and reasoning over external information such as news. While LLMs excel at language understanding and knowledge integration, they are not naturally designed for numerical prediction and time-series modeling. Bridging these two requirements remains a central challenge for financial foundation models.

Most modern forecasting models, from patch-based Transformers to large-scale time-series foundation models, are designed for signals that are smooth, strongly periodic, and governed by relatively stable dynamics, such as electricity load, traffic, and weather. Financial markets differ fundamentally: they are noisy, weakly autocorrelated, and characterized by rapidly changing volatility regimes. Moreover, financial time series are highly heterogeneous across assets, with volatility levels often spanning multiple orders of magnitude. As a result, representations learned under one scale regime frequently fail to generalize to others, while models trained on high-variance assets tend to overlook the subtle signals present in lower-variance markets. Bridging this gap requires a representation that captures transferable market patterns while remaining robust to large differences in scale and volatility.

A substantial fraction of price-moving information originates outside historical price trajectories. Earnings announcements, policy decisions, macroeconomic releases, and the collective sentiment reflected in financial news and social media continuously shape market behavior. Consequently, even highly expressive numerical forecasting models remain structurally blind to information that has already been disclosed through text. While recent large language models offer powerful language understanding and knowledge integration capabilities, they are not naturally optimized for numerical forecasting. Existing foundation models therefore tend to specialize in either numerical modeling or language reasoning, but rarely both. Closing this gap requires a bimodal forecasting framework that jointly reasons over historical market dynamics and contemporaneous textual information.

Large language models (LLMs) provide a promising foundation for addressing both challenges simultaneously: they already encode rich semantic knowledge and reasoning capabilities over text, while their autoregressive architecture offers a natural framework for sequential prediction. Recent efforts have adapted LLMs to time-series forecasting through three main paradigms: reprogramming continuous signals into the embedding space of a frozen LLM~\cite{timellm,gpt4ts,test}, converting numerical values into textual representations and leveraging next-token prediction directly~\cite{llmtime}, and quantizing time series into discrete tokens so that forecasting becomes language modeling over an extended vocabulary~\cite{chronos,totem}. For financial forecasting, we argue that the quantization paradigm is particularly appealing. By representing market dynamics as discrete tokens, it places price trajectories and textual information within a unified autoregressive sequence over a shared vocabulary, enabling joint modeling of historical prices and external news. More importantly, quantization transforms forecasting into a discrete prediction problem, producing verifiable targets that naturally support reinforcement learning with outcome-based rewards. Recent studies have shown that such verifiable rewards are effective for eliciting deliberate reasoning in forecasting tasks~\cite{tsfreasoning,counts}. In contrast, continuous reprogramming does not naturally provide an explicit discrete target space, while textual stringification may fragment numerical structure and incur substantial token overhead. These advantages make the quantize-and-adapt paradigm a natural foundation for building a bimodal financial forecasting model that integrates both numerical dynamics and textual reasoning.

Committing to the quantize-and-adapt paradigm turns the two challenges above into three design requirements. (1) Discretization must preserve scale information. Financial assets exhibit volatility levels spanning multiple orders of magnitude, and naive quantization either lets high-volatility assets dominate the codebook or collapses low-volatility series into near-degenerate representations. While normalization alleviates this issue, it also removes scale information that is often predictive. (2) Acquiring numerical forecasting ability must not come at the expense of textual reasoning. Fine-tuning an entire LLM on tokenized time series risks shifting model capacity toward numerical patterns and degrading the language understanding needed to interpret news and justify forecast revisions. (3) Fast forecasting and slow reasoning should coexist within a single model. Most predictions require only a lightweight history-to-future mapping, while text-conditioned deliberation is mainly valuable during regime shifts. A practical financial foundation model should therefore support both efficient forecasting and optional reasoning without maintaining separate systems or relearning the forecasting process.

We present \textsc{DualCast}, a dual-path framework built on a simple principle: a finite-capacity backbone should acquire numerical forecasting ability without sacrificing the pretrained linguistic and reasoning competence it later needs to interpret news. To preserve scale information across heterogeneous assets, the \emph{fast mode} employs a decoupled scale--shape tokenizer that represents normalized price dynamics and market scale as separate discrete tokens, enabling robust forecasting across diverse volatility regimes. These time-series tokens are introduced through a fully frozen Qwen3-8B backbone~\cite{qwen3}, with only the newly added token embeddings trained, thereby preserving the model's native language capabilities by construction. To support both efficient forecasting and optional deliberation, the \emph{slow mode} adds a toggleable adapter~\cite{lora} on the same frozen backbone. When enabled, it conditions on news and the fast-mode prediction to produce a revised forecast together with a reasoning trace, optimized via reinforcement learning in return space. Because both modes share the same backbone and forecasting format, reasoning can be invoked only when additional context is valuable, while retaining the efficiency of the fast forecasting path.

\begin{figure*}[t]
  \centering
  \includegraphics[width=0.98\textwidth]{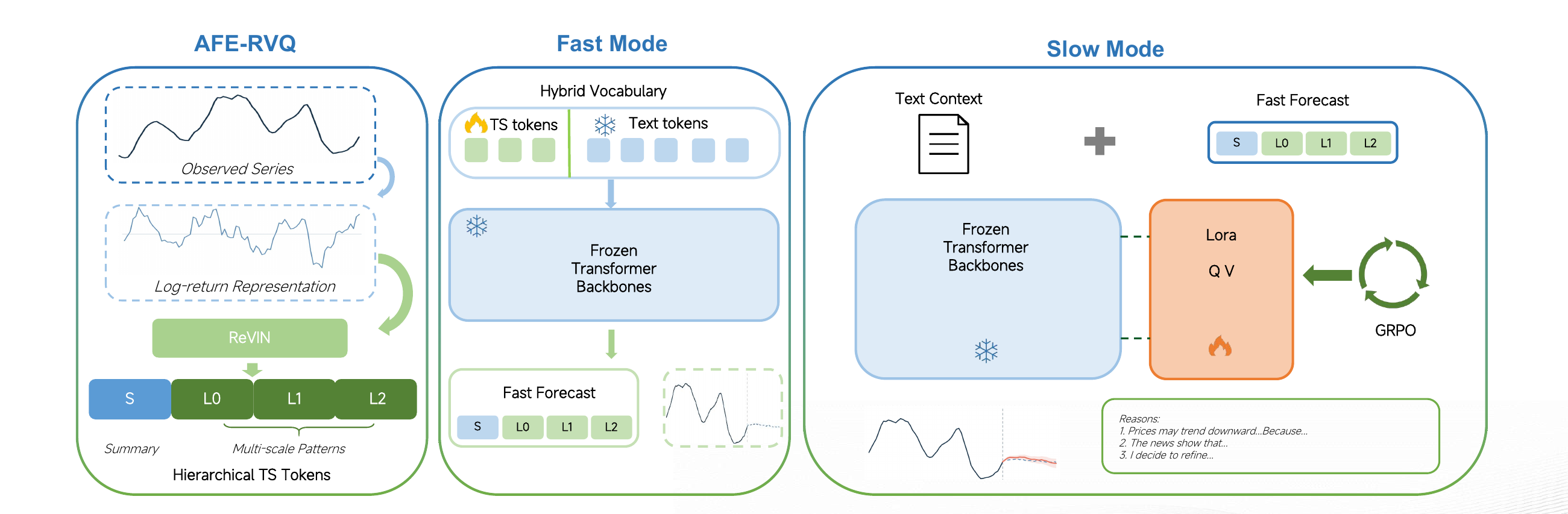}
  \caption{Overview of \textsc{DualCast}. The scale–shape tokenizer separates scale statistics from residual shape patterns and maps both to a hierarchical financial-token sequence. Fast mode predicts future financial tokens using the shared frozen backbone. Slow mode activates a LoRA adapter and uses causally available text together with the fast forecast to produce a news-conditioned revision. The adapter is initialized by supervised fine-tuning and subsequently optimized with the return-space GRPO objective described in Section~\ref{sec:method}.}
  \label{fig:overview}
\end{figure*}

Figure~\ref{fig:overview} summarizes the framework. The scale–shape tokenizer converts each return patch into one summary token and three residual shape tokens. Fast mode learns to forecast this financial-token sequence with a frozen language-model
backbone, whereas slow mode activates a lightweight adapter that conditions on
causally available text and the fast forecast to generate a revised path.

\paragraph{Contributions}
\begin{itemize}
  \item \textbf{Scale-aware financial tokenization.} We represent each log-return patch with one token from a learned summary codebook and three residual shape tokens. Adaptive frequency-equalizing residual vector quantization (AFE-RVQ) rebalances assignments within the shape codebooks, reducing codeword usage imbalance while maintaining reconstruction accuracy.
  \item \textbf{Parameter-efficient dual-path adaptation.} We extend a frozen Qwen3-8B backbone with financial tokens and train their embeddings and output heads for the fast forecasting path. A toggleable LoRA adapter enables a slow revision path on the same backbone. Across the two training stages, 48.9 million parameters, or 0.60\% of the model, are trainable.
  \item \textbf{News-conditioned revision with return-space feedback.} The slow path conditions on the fast forecast and news available at the forecast origin to produce a revised prediction. We initialize the reviser through supervised fine-tuning and further optimize it with group relative policy optimization, using return-space rewards for improvement over the fast forecast and cross-sectional consistency.
\end{itemize}

\section{Related Work}

\subsection{LLMs for Time-Series Forecasting}

Adapting language models to numerical forecasting has produced three
methodological families. \emph{Reprogramming} approaches keep the LLM
frozen and learn a thin interface that projects patched series into the
model's embedding space: Time-LLM~\cite{timellm} aligns patches with
text prototypes, GPT4TS~\cite{gpt4ts} fine-tunes only the normalization
and projection layers of a frozen GPT-2, and TEST~\cite{test} contrasts
time-series and text embeddings to activate the LLM. These methods
preserve the backbone but operate on continuous embeddings, which makes
them incompatible with the discrete, RL-optimizable targets we require.
\emph{Stringification} methods such as LLMTime~\cite{llmtime} render
numbers as digit strings and forecast by next-token sampling; they are
simple and training-free but waste sequence length on digits and
fracture numerical semantics. \emph{Quantization} methods discretize
the signal into a finite vocabulary and train a language model over it.
Chronos~\cite{chronos} bins scaled values into a fixed vocabulary,
while TOTEM~\cite{totem} learns a VQ-VAE codebook of temporal patterns;
both demonstrate that ``time series as language'' transfers the mature
machinery of sequence modelling to forecasting. In parallel, a line of
domain-specific foundation models---TimesFM~\cite{timesfm},
Moirai~\cite{moirai}, MOMENT~\cite{moment}, Timer~\cite{timer},
Time-MoE~\cite{timemoe}, and Sundial~\cite{liu2025sundial}---pretrains
Transformers from scratch on large numerical corpora. KTD-Fin\cite{zhu2026knowing} finds no persistent positive stock-selection alpha among the LLM trading agents evaluated under leakage-controlled conditions. These models are
strong unimodal forecasters but discard the pretrained text and
reasoning priors that bimodal financial forecasting depends on.

Our tokenizer sits in the quantization family but departs from it in
two ways. First, following the patch-wise consensus of
PatchTST~\cite{patchtst} and PITS~\cite{pits} and per-instance
normalization of RevIN~\cite{revin}, we explicitly \emph{factor} each
patch into a normalized shape and a removed scale, then re-quantize the
scale with a learned codebook rather than discarding it or binning it
coarsely. Second, we adopt residual vector quantization from neural
audio coding~\cite{soundstream} with the rotation-trick straight-through
estimator~\cite{rotationtrick} and EMA codebooks~\cite{vqvae}, and add a
frequency-equalization step so the resulting LLM vocabulary is densely
utilized.

\subsection{Text--Time-Series Fusion for Financial Forecasting}

Combining textual signals with market data has a long history.
StockNet~\cite{stocknet} pairs tweets with prices for movement
prediction and remains a standard benchmark; FinBERT~\cite{finbert}
adapts BERT to financial sentiment; and BloombergGPT~\cite{bloomberggpt}
and FinGPT~\cite{fingpt} build finance-specialized LLMs, though their
temporal modelling is weak. A second thread designs explicit fusion
architectures---attention gating, cross-attention between text and
price, and contrastive alignment---typified by market-guided
transformers such as MASTER~\cite{master}. These systems generally
treat text and series as two encoders fused by a bespoke module, and
the numerical channel is a regression head rather than a generative
sequence. We instead place price tokens and news tokens in a
\emph{single} autoregressive sequence over a shared vocabulary, so
fusion is performed by the backbone's native attention and the model
can both read news and \emph{generate} a refined numerical forecast.

\subsection{Reasoning and Reinforcement Learning for Time Series}

A recent direction reframes forecasting as explicit reasoning rather
than a one-shot mapping. ``Slow-thinking'' studies~\cite{tsfreasoning,
slowthinking} show that prompting an LLM to deliberate before
predicting can outperform direct prediction, and surveys document a
rapid shift toward reasoning- and agent-centric time-series
systems~\cite{tsreasoningsurvey}; reinforcement learning over discrete
token outputs has emerged as the standard way to elicit such
chain-of-thought via verifiable rewards~\cite{counts}. We follow this
philosophy---reasoning is \emph{elicited by RL with verifiable
rewards}~\cite{grpo,dapo} rather than distilled from synthetic
traces---but our contribution is orthogonal to how the reasoning trace
itself is produced, and lies instead in \emph{where} and \emph{how
cheaply} reasoning is added. First, we keep the backbone permanently
frozen and adapt only a compact set of new embeddings, so the model's
native text and reasoning ability is preserved for the refinement stage
instead of being diluted by domain fine-tuning. Second, reasoning is an
\emph{optional} second pass: a single toggleable adapter recovers the
fast single-forward forecaster when disabled and a slow, text-conditioned
reasoner when enabled. Third, the refinement sequence regenerates the
forecast in a numerical context, so the history-to-target attention
learned during pretraining is reused at refinement time rather than
relearned. Together these make the reasoning path inexpensive to add and
inexpensive to skip, which is essential when a fast estimate already
suffices and deliberation is warranted only at regime shifts.

\section{Method}
\label{sec:method}

\subsection*{Overview and Problem Setup}

Let $p_{1:C+H}=\{p_1,\ldots,p_{C+H}\}$ denote a univariate
positive price series, where $p_{1:C}$ is the historical context and
$p_{C+1:C+H}$ is the forecasting target. In the multimodal setting, the
forecast origin is also associated with a timestamped text stream
$\mathcal{T}_{\le C}$, such as news headlines, filings, or social media
posts, whose timestamps are no later than the forecast origin. The task
is to predict the future price path while respecting this no-leakage
constraint.

\textsc{DualCast} decomposes the task into a fast numerical path and a
slow text-conditioned refinement path. Stage~1 learns a time-series
language: prices are converted into discrete tokens and a frozen LLM is
trained to autoregressively forecast the next tokens. Stage~2 keeps the
same frozen backbone and Stage~1 time-series vocabulary, attaches a
small adapter, and asks the model to revise the Stage~1 forecast using
recent text. This design separates two capabilities that are both
needed in financial forecasting: robust pattern matching over noisy
price histories and language-based reasoning over exogenous events.

\subsection{Shape--Scale Tokenization with AFE-RVQ}
\label{subsec:tok}

A local price segment contains two fundamentally different types of
information: its movement pattern and its scale. The former describes
relative temporal dynamics, such as rallies, reversals, or
consolidation behaviors, while the latter reflects local drift and
volatility. Quantizing both factors within a single representation space
forces the tokenizer to model heterogeneous phenomena simultaneously and
reduces token efficiency. We therefore decompose each patch into a
\emph{shape} component and a \emph{scale} component, which are
discretized separately and later combined into a unified token stream.

We represent the price trajectory by log-returns rather than raw
prices,
\begin{equation}
r_t=\log p_t-\log p_{t-1},
\end{equation}
which removes dependence on the absolute price level and makes the
representation invariant to multiplicative rescaling. The return series
is partitioned into non-overlapping patches of length $L$, and each
patch is independently normalized using reversible instance
normalization (RevIN)~\cite{revin}. The normalized patch serves as the
shape signal, while the removed normalization statistics are retained as
patch-level scale information.

To bridge continuous financial signals and discrete language modeling,
we convert each normalized patch into a small set of discrete shape
tokens. An encoder first maps the normalized patch
$\tilde r^{(w)}$ into a latent representation, which is then
discretized using residual vector quantization (RVQ)~\cite{soundstream}.
By representing each patch as a composition of codewords drawn from
multiple codebooks, RVQ provides a compact yet expressive vocabulary of recurring return patterns. The resulting code indices form the shape
tokens, allowing numerically different but structurally similar price
movements to share a common symbolic representation.

Because every RVQ codeword becomes one trainable embedding row in the
extended LLM vocabulary, its assignment frequency directly determines how
often that row receives a gradient. Plain RVQ can concentrate a disproportionate
fraction of assignments on a small number of codewords, producing a peaked
financial-token distribution even when most entries are active. We therefore
introduce \emph{Adaptive Frequency-Equalizing RVQ} (AFE-RVQ), a periodic
rebalancing operator applied independently at each residual level during
tokenizer training.

Let $\mathcal{C}=\{c_i\}_{i=1}^{K}$ be one codebook and let $h_i$ denote the
number of latent vectors assigned to $c_i$ over a pool $\mathcal{Z}$ of recent
batches. With the uniform expected load $\bar h=|\mathcal{Z}|/K$, AFE-RVQ
identifies overloaded and starved entries as
\begin{equation}
\mathcal{O}=\{i:h_i>\tau_{\mathrm{hi}}\bar h\},\qquad
\mathcal{D}=\{i:h_i<\tau_{\mathrm{lo}}\bar h\}.
\end{equation}
For an overloaded entry $i$, the latent vectors assigned to $c_i$ are split
with a small $k$-means problem. One sub-center replaces $c_i$, while the
remaining sub-centers are written into available entries in $\mathcal{D}$.
The exponential-moving-average accumulators of every relocated entry are
reset so that it is not pulled back toward its previous location, and a small
anti-symmetric perturbation separates sibling sub-centers that would otherwise
re-merge. In our deployed tokenizer, assignment counts are collected from
$P=64$ recent batches and rebalancing is applied every 500 steps with
$\tau_{\mathrm{lo}}=0.5$ and $\tau_{\mathrm{hi}}=2.0$. The operation adds no
parameters and is disabled at inference time. As shown in
Section~\ref{sec:afe}, it reduces peak codeword load while preserving
reconstruction fidelity; we therefore use it to obtain a better-balanced
vocabulary rather than claim additional representational capacity.

Since the normalization used to isolate shape patterns removes local drift and
volatility, the discarded scale information is represented separately using a
discrete summary token.

The retained statistics $(\mu_w,\sigma_w)$ are discretized using a
learned summary vocabulary constructed from observed
$(\mu_w,\log\sigma_w)$ pairs in the training set. Each summary token
corresponds to a frequently occurring local market regime characterized
by a particular combination of drift and volatility. Unlike manually
designed discretization grids, the learned vocabulary allocates capacity
according to the empirical distribution of market states and therefore
uses tokens more efficiently.

Each patch is ultimately represented by one summary token $S$ followed by
three residual shape tokens,
\begin{equation}
[S,L_0,L_1,L_2].
\end{equation}
Placing the summary token first allows subsequent shape tokens to be
interpreted under the corresponding local market regime. The resulting
token sequence forms a discrete financial language that can be modeled
directly by a standard autoregressive LLM.

\subsection{Fast Mode: Financial Language Modeling}
\label{subsec:fast}

Fast mode treats the tokenized price sequence as a discrete financial
language and trains a frozen LLM to autoregressively model its dynamics.
Given a sequence of summary and shape tokens produced by AFE-RVQ, the
model predicts future tokens using only historical price information.
This mode serves as the efficient forecasting path of \textsc{DualCast}
and provides the baseline prediction that is later refined by the slow
mode.

We extend the Qwen3-8B vocabulary with the financial tokens introduced
in Section~\ref{subsec:tok}, including one summary vocabulary and three
shape vocabularies. Rather than random initialization, the newly added
embeddings inherit the geometry of the learned tokenizers, preserving
similarity relationships among both market regimes and local price
patterns. Throughout fast-mode training, the pretrained LLM backbone
remains frozen. The resulting token stream follows a deterministic
grammar that uniquely determines the type of the next token from its
position, enabling a factorized prediction objective in which the
softmax is restricted to the valid token subset associated with the next token type rather than competing over the entire vocabulary. This design eliminates interference from irrelevant text tokens and better aligns
optimization with the structure of the discrete financial language.

The overall training objective is
\begin{equation}
\mathcal{L}_{\mathrm{fast}}
=
\lambda_{\textsc{sum}} \, \mathcal{L}_{\textsc{sum}}^{\mathrm{CE}}
+
\sum_{k=0}^{2} \lambda_{L_k} \, \mathcal{L}_{L_k}^{\mathrm{CE}}.
\end{equation}
where later residual levels receive smaller weights due to their higher entropy and lower predictability. The first patch of each sequence is
used only as context, while all subsequent tokens contribute
autoregressive prediction losses.

Fast mode learns an interface between the discrete financial language
and the pretrained LLM. Only the newly introduced financial-token
embeddings and prediction heads are optimized, while all original Qwen
parameters remain fixed. This design allows the model to acquire
forecasting ability without altering its native language and reasoning
capabilities, enabling the same backbone to be reused by both the fast
and slow modes of \textsc{DualCast}.

\subsection{Slow Mode: Text-Conditioned Forecast Revision}

While the fast mode captures numerical regularities from historical prices, it is intentionally blind to textual information. The slow mode therefore formulates multimodal forecasting as a \emph{forecast revision} problem, refining a strong fast-mode baseline using news and recent market context rather than generating a forecast from scratch. Such a revision-based formulation mirrors how investors typically reason about markets: they first extrapolate from observed trends and then revise their expectations in response to newly arriving information.

Formally, given historical prices $p_{1:C}$, related news
$\mathcal{T}_{\le C}$, and a fast-mode forecast
$\mathbf{y}_{\mathrm{fast}}$, the slow mode produces a refined forecast
$\mathbf{y}_{\mathrm{slow}}$ by conditioning on all three sources of
information.

To realize this revision process, we express each example in the native chat format of the base language model. The prompt contains three explicitly separated information sources: historical prices, related news, and the fast-mode forecast, while the model is instructed to generate a refined forecast for the target horizon together with a reasoning trace.

The slow mode is first initialized through supervised fine-tuning on automatically constructed revision examples. For each price--news pair, the base model is prompted to analyze the potential market implications of the news and produce a rationale followed by a refined forecast. To better align textual reasoning with the newly introduced time-series vocabulary, we augment a subset of training examples with an explicit trend description derived from recent price movements at the beginning of the reasoning trace. This lightweight grounding procedure encourages the model to associate time-series tokens with semantic concepts such as upward, downward, or sideways market behavior.

Supervision is applied primarily to the refined forecast tokens, while prompt tokens are masked. The trend-grounding examples serve only as an auxiliary alignment signal and do not alter the forecasting target. As a result, the model learns both the revision behavior and the structured output format without requiring manually annotated financial reasoning traces.

\subsection{Slow Mode: Utility Optimization via Revision Feedback}

While supervised fine-tuning enables the model to imitate revision
behavior from automatically constructed examples, it primarily optimizes
token-level likelihood and does not directly reflect forecasting utility.
In particular, it does not explicitly encourage the model to improve
upon the fast-mode baseline in a decision-relevant manner. As a result,
the model may produce plausible revisions without consistently improving
predictive performance.

To address this limitation, we optimize the slow mode using
group-relative policy optimization (GRPO), directly operating in the
forecast (return) space. The training objective is defined as a
composite reward:
\begin{equation}
R
=
w_{\mathrm{imp}} R_{\mathrm{imp}}
+
w_{\mathrm{IC}} R_{\mathrm{IC}}
+
w_{\mathrm{fmt}} R_{\mathrm{fmt}},
\end{equation}
where $R_{\mathrm{imp}}$ measures improvement over the fast-mode
baseline, $R_{\mathrm{IC}}$ captures cross-sectional directional
consistency, and $R_{\mathrm{fmt}}$ enforces valid structured output.

Let $\rho(x)$ denote the cumulative return implied by forecast $x$, and
let $\rho^\star$ denote the realized cumulative return. We define a
volatility-normalized error:
\begin{equation}
e(x) = \frac{|\rho(x) - \rho^\star|}{s},
\qquad
s = 2\sigma_C \sqrt{H},
\end{equation}
where $\sigma_C$ is estimated from the historical context and $H$ is the
forecast horizon. The improvement reward is then defined as:
\begin{equation}
R_{\mathrm{imp}} =
\operatorname{clip}\!\left(
\frac{e(\mathbf{y}_{\mathrm{fast}}) - e(\mathbf{y}_{\mathrm{slow}})}
{\max(e(\mathbf{y}_{\mathrm{fast}}), \Delta)},
-1, 1
\right).
\end{equation}

This formulation ensures that the slow mode is rewarded only when it
strictly improves upon the fast-mode forecast, explicitly aligning
optimization with the goal of forecast revision rather than standalone
prediction.

While $R_{\mathrm{imp}}$ measures absolute improvement, financial
decisions are often driven by correct cross-sectional ranking rather
than point-wise accuracy. We therefore introduce a rank-based
directional reward computed over a rollout batch.

For each sample $i$ in a batch, we compute volatility-normalized returns:
\begin{equation}
z_i = \frac{\rho(\mathbf{y}_i)}{2\sigma_i \sqrt{H_i}},
\qquad
z_i^\star = \frac{\rho_i^\star}{2\sigma_i \sqrt{H_i}}.
\end{equation}

We define a normalized rank operator $\mathcal{R}(\cdot)$ that maps
values within a batch to $[0,1]$ with tie averaging:
\begin{equation}
\tilde{z}_i = \mathcal{R}(z_i), \qquad
\tilde{z}_i^\star = \mathcal{R}(z_i^\star).
\end{equation}

We define a cross-sectional rank-consistency reward using the normalized within-batch ranks of the predicted and realized volatility-normalized returns:
\begin{equation}
R_{\mathrm{IC},i} =
4\left(\tilde{z}_i - \frac{1}{2}\right)
\left(\tilde{z}_i^\star - \frac{1}{2}\right),
\end{equation}
Here, ($\tilde z_i$) and ($\tilde z_i^\star$) denote the predicted and realized ranks, respectively. The reward is positive when the two ranks fall on the same side of the rank midpoint and negative when they fall on opposite sides. Aggregated over a rollout batch, it encourages agreement in the relative ordering of returns across assets. Because it depends only on ranks, this reward does not directly measure the sign or numerical magnitude of an individual predicted return; forecast accuracy is addressed separately by the return-space improvement reward.

We include a lightweight format reward $R_{\mathrm{fmt}}$ to enforce
valid structured generation, including separation between reasoning
traces and forecast tokens. It additionally penalizes language collapse
in the reasoning trace, discouraging degenerate repetition and the
leakage of time-series tokens into the textual rationale. This term
primarily acts as a constraint and contributes little gradient once
training stabilizes.

The reinforcement learning objective aligns the slow mode with the goal
of forecast revision: improving upon a strong numerical baseline while
remaining sensitive to cross-sectional market structure. This ensures
that optimization is driven by forecasting utility rather than language
modeling likelihood.

\section{Experiments}
\label{sec:exp}

We evaluate \textsc{DualCast} on financial forecasting across multiple markets, sampling frequencies, and modalities to assess both its forecasting accuracy and its ability to generalize beyond the distributions seen during pretraining. We first compare \textsc{DualCast} against strong time-series foundation models, multimodal forecasters, and general-purpose LLMs under a strict zero-shot protocol. We then investigate its ability to transfer across heterogeneous assets and unseen temporal resolutions, and finally analyze the contribution of the proposed scale--shape tokenizer and deliberative revision mechanism.

\subsection{Experimental Setup}

\textbf{\paragraph{Datasets}}
We evaluate on three benchmark sources covering two financial markets, three
temporal resolutions, and both price-only and multimodal forecasting settings.
Together, they assess generalization across markets, sampling frequencies, and
input modalities. \textbf{MTBench-short}~\cite{chen2025mtbench} provides event-aligned
five-minute U.S.\ equity windows. \textbf{FinMultiTime}~\cite{xu2025finmultitime}
provides daily OHLC price series paired with timestamped financial news for U.S.\
equities (\textsc{SP500}) and Chinese equities (\textsc{HS300}). We use only text
available at the forecast origin, ensuring strict temporal causality. Table~\ref{tab:data} summarizes the evaluation datasets, markets,
temporal resolutions, and input modalities. \textbf{Time-MMD (Energy)}~\cite{liu2024time} consists of weekly U.S.\ gasoline
retail prices paired with textual market reports, extending evaluation
beyond equity markets to commodity price forecasting.

\textbf{\paragraph{Window construction}}
All settings use a lookback of $C{=}96$ observations and horizons
$H\!\in\!\{16,32,64\}$. We construct the evaluation origins using the
maximum horizon $H_{\max}{=}64$ and then reuse exactly the same origins for the
shorter horizons by truncating the target to its first 16 or 32 observations. All baselines follow the same construction. Table~\ref{tab:params} summarizes the parameter-efficient training footprint of \textsc{DualCast}; the backbone remains frozen throughout
both training stages.


\begin{table}[t]
\centering
\caption{Evaluation datasets, covering markets, sampling frequencies, and input modalities. All datasets are evaluated zero-shot at horizons \(H\in\{16,32,64\}\).}
\label{tab:data}
\footnotesize
\setlength{\tabcolsep}{4pt}
\begin{tabular}{llc}
\toprule
Dataset & Market / freq. & Modality \\
\midrule
MTBench-short~\cite{chen2025mtbench}      & US eq., 5-min  & price only \\
\textsc{SP500}~\cite{xu2025finmultitime}  & US eq., daily   & price\,+\,news \\
\textsc{HS300}~\cite{xu2025finmultitime}  & CN eq., daily   & price\,+\,news \\
Energy~\cite{liu2024time}       & US gas, weekly  & price\,+\,news \\
\bottomrule
\end{tabular}
\end{table}

\vspace{5pt}

\begin{table}[t]
\centering
\caption{Parameter-efficient training footprint of \textsc{DualCast}. The 8.2B backbone remains frozen; 48.9M parameters (0.60\%) are trained across the two stages.}
\label{tab:params}
\footnotesize
\setlength{\tabcolsep}{4pt}
\begin{tabular}{lrl}
\toprule
Component & Params & Trained \\
\midrule
Qwen3-8B backbone                       & $8.19$\,B & frozen \\
\;+\,Financial tokens ($4{,}097$; embed\,+\,head) & $33.6$\,M & Stage~1 \\
\;+\,LoRA adapter ($q,v$; $r{=}32$)     & $15.3$\,M & Stage~2 \\
\midrule
Total trainable                         & $48.9$\,M & $\mathbf{0.60\%}$ \\
\bottomrule
\end{tabular}
\end{table}

\begin{table*}[!t]
\centering
\caption{Forecasting MAPE (\%, $\downarrow$) across markets and frequencies.
MTBench-short is five-minute and evaluated with price only;
\textsc{SP500}/\textsc{HS300} are daily, and Energy is weekly.
\textsc{DualCast} uses no dataset-specific fine-tuning. Best per column in
\textbf{bold}.}
\label{tab:main}
\footnotesize
\setlength{\tabcolsep}{3.2pt}
\begin{tabular}{lcccccccccccc}
\toprule
& \multicolumn{3}{c}{MTBench (5-min)} & \multicolumn{3}{c}{\textsc{SP500}} & \multicolumn{3}{c}{\textsc{HS300}} & \multicolumn{3}{c}{Energy} \\
\cmidrule(lr){2-4}\cmidrule(lr){5-7}\cmidrule(lr){8-10}\cmidrule(lr){11-13}
Model & $16$ & $32$ & $64$ & $16$ & $32$ & $64$ & $16$ & $32$ & $64$ & $16$ & $32$ & $64$ \\
\midrule
Chronos-2   & 1.51 & 1.77 & 2.07 & 4.98 & 6.82 & 11.12 & \textbf{5.84} & \textbf{7.13} & 9.50 & 7.11 & 10.47 & 13.56 \\
TimesFM-2.5 & 1.47 & 1.73 & 2.04 & 5.03 & 7.01 & 11.25 & 6.10 & 7.21 & 9.74 & \textbf{7.09} & 10.43 & 13.65 \\
Timer       & 1.59 & 1.90 & 2.25 & 7.04 & 9.61 & 15.17 & 6.38 & 7.77 & 10.63 & 7.76 & 10.70 & 13.91 \\
Time-MoE    & 1.57 & 1.90 & 2.31 & 6.55 & 9.05 & 14.23 & 6.06 & 8.25 & 11.09 & 7.45 & 10.62 & 14.05 \\
Moirai-2.0  & 1.46 & 1.72 & 2.03 & 5.03 & 7.09 & 11.29 & 6.17 & 7.20 & 9.49 & 7.13 & \textbf{10.20} & 13.59 \\
ChatTime  & 2.10 & 2.41 & 2.72 & 5.68 & 7.43 & 11.39 & 6.82 & 7.91 & 10.14 & 15.55 & 16.49 & 17.60 \\
Aurora    & 3.34 & 3.47 & 3.60 & 12.68 & 14.20 & 18.20 & 8.86 & 9.78 & 12.52 & 13.59 & 14.79 & 16.72 \\
Qwen3 & 4.50 & 4.75 & 5.60 & 25.80 & 25.66 & 31.15 & 15.64 & 17.37 & 22.51 & 94.46 & 57.06 & 40.87 \\
\midrule
\textsc{DualCast} (Slow) & \textbf{1.45} & \textbf{1.68} & \textbf{1.96} & \textbf{4.81} & \textbf{6.45} & \textbf{9.80} & 7.08 & 7.58 & \textbf{9.19} & 7.38 & 10.33 & \textbf{13.30} \\
\bottomrule
\end{tabular}
\end{table*}

\textbf{\paragraph{Baselines}}
We compare against three families: (1) \emph{time-series foundation models}, in
their latest released versions, Chronos-2~\cite{ansari2025chronos},
TimesFM-2.5~\cite{timesfm}, Timer \cite{timer}, and
Time-MoE~\cite{timemoe}, and Moirai-2.0~\cite{moirai}; (2)
\emph{dedicated multimodal forecasters} that natively consume text,
ChatTime~\cite{wang2025chattime} and Aurora~\cite{wu2026aurora}; and (3) an
LLM-prompting baseline that feeds the same recent prices and news to a
general-purpose Qwen3-8B~\cite{qwen3} as text and parses its numeric forecast. All baselines run in their released zero-shot configurations on the same windows and
horizons. 

\textbf{\paragraph{Metrics}}
Because financial series span several orders of magnitude in scale and absolute
errors are not comparable across assets, we report mean absolute percentage error
(\textbf{MAPE}, \%) at horizons $H\!\in\!\{16,32,64\}$. For \textsc{DualCast} we
report the fast forecaster (\textsc{Fast}), the SFT reviser (\textsc{SFT}), and
the RL-tuned slow reviser (\textsc{Slow}); \textsc{Slow} is the default.

\subsection{Main Results}
DualCast (Slow) achieves the lowest MAPE in 8 of 12 dataset–horizon settings as Table~\ref{tab:main} shows, including H=64 on all four evaluation subsets. On five‑minute MTBench, whose resolution is entirely unseen during pretraining, \textsc{Slow} obtains the best results across all horizons. This demonstrates that the scale–shape tokenizer is able to extract market patterns that are agnostic to sampling frequency. Despite keeping the entire backbone frozen, our method achieves competitive or superior performance compared with cutting-edge time-series foundation models, including Chronos-2, TimesFM-2.5, and Moirai-2.0, especially at long forecasting horizons critical for real-world financial analysis. Consistent with baseline protocols, DualCast follows the same price-only pretraining paradigm and only updates merely \(0.6\%\) of the total parameters. \textsc{DualCast} clearly outperforms ChatTime and Aurora; feeding long news text into ChatTime even \emph{inflates} its error on Energy ($\geq\!15\%$), as serializing text disrupts its numeric scale anchoring, whereas our method keeps the model anchored to the price dynamics. Figures~\ref{fig:forecast}
and~\ref{fig:modeexamples} visualize stable forecasts across markets and the Fast--SFT--Slow progression on identical windows, respectively.

\subsection{Refinement: Fast vs.\ Slow}
\label{sec:ablation}

\begin{figure*}[!t]
\centering
\begin{minipage}{0.49\textwidth}\centering
  \includegraphics[width=\linewidth]{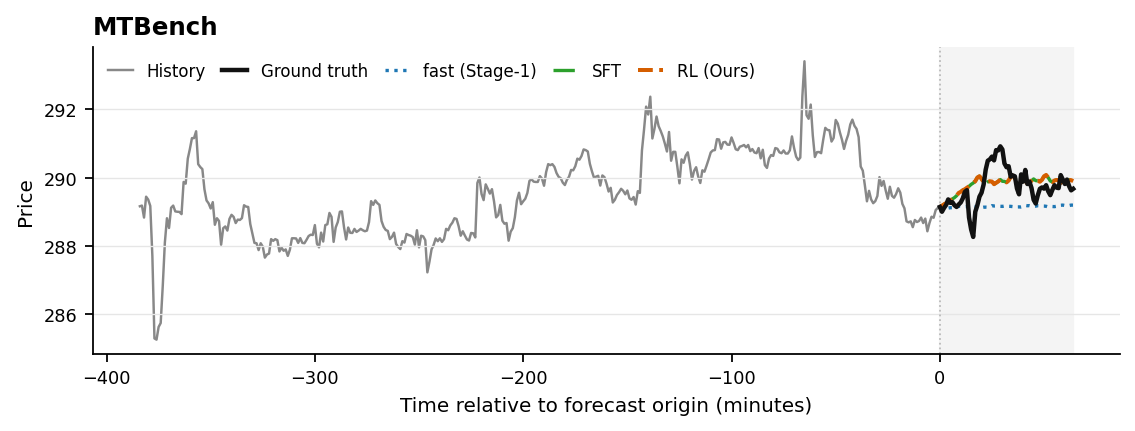}\\[-2pt]
  {\small (a) MTBench (5-minute)}
\end{minipage}\hfill
\begin{minipage}{0.49\textwidth}\centering
  \includegraphics[width=\linewidth]{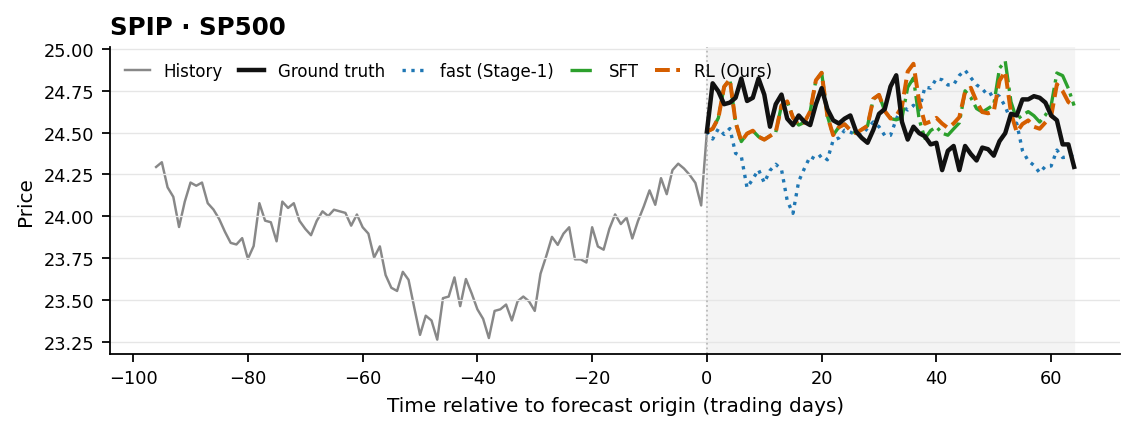}\\[-2pt]
  {\small (b) SPIP, \textsc{SP500} (daily)}
\end{minipage}
\caption{Representative Fast--SFT--Slow forecasts on identical windows. Gray
denotes the observed history, black the realized future, blue the fast Stage-1
forecast, green the SFT reviser, and orange the RL-tuned slow reviser. The
MTBench example shows a small refinement when the fast estimate is already
reasonable; the SP500 example shows the two refinement stages correcting a
larger fast-mode deviation. These cases are illustrative; aggregate improvements
are reported in Table~\ref{tab:ablation}.}
\label{fig:modeexamples}
\end{figure*}

\textsc{DualCast} provides three forecasting modes based on the same frozen backbone: the fast numerical forecaster (Fast), a supervised fine-tuned reviser (SFT), and a reviser further optimized by reinforcement learning (Slow). Table 4 compares these modes on identical evaluation windows. MAPE decreases or remains unchanged from Fast to SFT to Slow in every dataset–horizon setting; SFT and Slow tie on MTBench-short at H=16 and H=64. At H=64, MAPE decreases from 18.06\% to 12.43\% and then 9.19\% on HS300, and from 15.26\% to 14.03\% and then 9.80\% on SP500. These comparisons show that revision improves the fast forecast across the evaluated settings, with larger gains at longer horizons. Figure~\ref{fig:modeexamples} illustrates this progression on representative windows.

\begin{table}[!t]
\centering
\caption{Ablation of the three \textsc{DualCast} modes on identical windows.
MAPE (\%, $\downarrow$); best per row group in \textbf{bold}. MTBench-short is
five-minute and price-only; \textsc{SP500}/\textsc{HS300} are daily, and Energy
is weekly.}
\label{tab:ablation}
\small
\setlength{\tabcolsep}{4pt}
\begin{tabular}{llccc}
\toprule
Dataset & Mode & $H{=}16$ & $32$ & $64$ \\
\midrule
\multirow{3}{*}{MTBench}
 & Fast & 1.49 & 1.76 & 2.05 \\
 & SFT  & \textbf{1.45} & 1.69 & \textbf{1.96} \\
 & Slow & \textbf{1.45} & \textbf{1.68} & \textbf{1.96} \\
\midrule
\multirow{3}{*}{\textsc{SP500}}
 & Fast & 6.23 & 9.03 & 15.26 \\
 & SFT  & 5.68 & 7.87 & 14.03 \\
 & Slow & \textbf{4.81} & \textbf{6.45} & \textbf{9.80} \\
\midrule
\multirow{3}{*}{\textsc{HS300}}
 & Fast & 8.75 & 11.61 & 18.06 \\
 & SFT  & 7.56 & 8.79 & 12.43 \\
 & Slow & \textbf{7.08} & \textbf{7.58} & \textbf{9.19} \\
\midrule
\multirow{3}{*}{Energy}
 & Fast & 9.73 & 15.02 & 24.48 \\
 & SFT  & 7.73 & 11.08 & 14.24 \\
 & Slow & \textbf{7.38} & \textbf{10.33} & \textbf{13.30} \\
\bottomrule
\end{tabular}
\end{table}

\begin{table*}[!t]
\centering
\caption{News ablation: point error (MAPE, \%, $\downarrow$) and directional
accuracy ($\uparrow$) of the \textsc{Slow} reviser with real news vs.\ with news
removed, on identical date-aligned windows. Better of
the two conditions per column in \textbf{bold}.}
\label{tab:newsabl}
\small
\setlength{\tabcolsep}{6pt}
\begin{tabular}{ll ccc ccc}
\toprule
 & & \multicolumn{3}{c}{MAPE (\%, $\downarrow$)} & \multicolumn{3}{c}{Directional acc.\ ($\uparrow$)} \\
\cmidrule(lr){3-5}\cmidrule(lr){6-8}
Dataset & News & $H{=}16$ & $32$ & $64$ & $H{=}16$ & $32$ & $64$ \\
\midrule
\multirow{2}{*}{Energy}
 & removed & \textbf{7.30} & 10.37 & 13.48 & 0.49 & 0.42 & 0.45 \\
 & real    & 7.31 & \textbf{10.32} & \textbf{13.27} & \textbf{0.55} & \textbf{0.54} & \textbf{0.51} \\
\midrule
\multirow{2}{*}{\textsc{SP500}}
 & removed & 5.67 & 7.55 & 10.64 & 0.51 & 0.52 & 0.48 \\
 & real    & \textbf{5.63} & \textbf{7.49} & \textbf{10.56} & \textbf{0.53} & \textbf{0.55} & \textbf{0.50} \\
\midrule
\multirow{2}{*}{\textsc{HS300}}
 & removed & 6.41 & 7.91 & 10.04 & \textbf{0.53} & \textbf{0.57} & 0.54 \\
 & real    & \textbf{6.25} & \textbf{7.68} & \textbf{9.77} & 0.48 & 0.55 & \textbf{0.55} \\
\bottomrule
\end{tabular}
\end{table*}

Table~\ref{tab:main} also shows that directly prompting Qwen3-8B with recent prices and news as text yields substantially higher MAPE than \textsc{DualCast} (Slow) on the daily equity subsets. On SP500, Qwen3-8B records MAPE of 25.66–31.15\% across the three horizons, compared with 4.81–9.80\% for \textsc{DualCast} (Slow). On HS300, the corresponding ranges are 15.64–22.51\% and 7.08–9.19\%. Thus, direct text prompting is a weaker baseline in these evaluations.

\begin{table}[!t]
\centering
\caption{Cost of deliberation: per-forecast latency (median) and number of
generated tokens on \textsc{SP500} (single H800, batch~1, greedy decoding).
The fast path decodes only the forecast tokens; the slow path
additionally generates a reasoning trace and regenerates the forecast.}
\label{tab:cost}
\small
\begin{tabular}{lcc}
\toprule
Mode & Latency (s) & Avg Generated tokens \\
\midrule
\textsc{Fast} & 0.37 & 16 \\
\textsc{Slow} & 5.85 & $199$ \\
\bottomrule
\end{tabular}
\end{table}

Table~\ref{tab:cost} quantifies the price of reasoning on \textsc{SP500}
(single H800, batch~1, greedy decoding). The fast path emits only the
$16$ forecast tokens and completes in $0.37$\,s per window, whereas the slow path
autoregressively generates a $\sim\!167$-token reasoning trace before regenerating
the forecast, costing $5.85$\,s---about $16\times$ more. Because the adapter is
toggleable and both modes share the frozen backbone, disabling deliberation
recovers the fast forecaster exactly, with no separate model to maintain. The
reasoning pass is therefore something to spend only when text is likely to matter, while the large majority of forecasts are served by the cheap fast path.

\subsection{Does Text Help? A News Ablation}
\label{sec:newsabl}

The revision-based slow mode is designed to fold textual information into a
numerical forecast, but MAPE alone cannot reveal whether the text is actually
used: a lower error may simply reflect the RL stage sharpening the numerical
baseline. To isolate the contribution of text, we re-run the \textsc{Slow}
reviser on strictly date-aligned windows under two conditions--with the real
news at the forecast origin, and with the news removed--holding everything else
fixed. To assess whether news affects forecast direction as well as point accuracy, we also report directional accuracy.

Table~\ref{tab:newsabl} shows that the text channel is useful, and we evaluate performance via two complementary metrics: point‑prediction MAPE and directional accuracy. For \textbf{point
error}, real news lowers MAPE in eight of the nine dataset--horizon comparisons.
The gains are largest and most consistent on \textsc{HS300}
($-0.16/-0.23/-0.27$), remain positive on \textsc{SP500}
($-0.04/-0.06/-0.08$), and appear at the longer Energy horizons
($-0.05/-0.21$ for $H{=}32/64$); the only exception is a negligible $+0.01$
increase on Energy at $H{=}16$. For \textbf{direction}, Energy improves at every
horizon ($+0.06/+0.12/+0.06$), and \textsc{SP500} shows smaller consistent gains
($+0.02/+0.03/+0.02$). \textsc{HS300} is mixed: news improves point accuracy at
all horizons but directional accuracy only at $H{=}64$. These ablations indicate that real news improves point accuracy in most tested settings, while its effect varies across datasets, horizons, and metrics. They also show that news conditioning is not uniformly beneficial for every metric and horizon, motivating selective activation of the
slow path rather than unconditional deliberation.

\begin{figure*}[!t]
\centering
\begin{minipage}{0.49\textwidth}\centering
  \includegraphics[width=\linewidth]{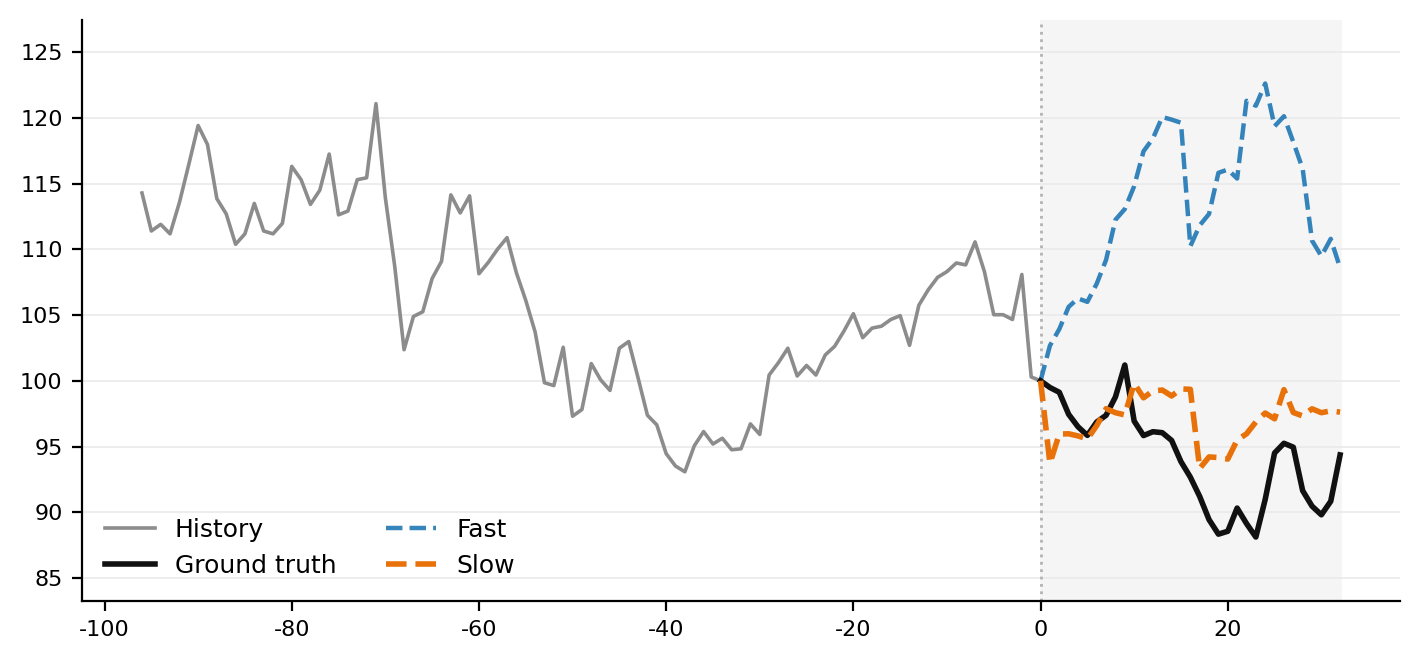}\\[-2pt]
  {\small (a) HPE, $H{=}32$}
\end{minipage}\hfill
\begin{minipage}{0.49\textwidth}\centering
  \includegraphics[width=\linewidth]{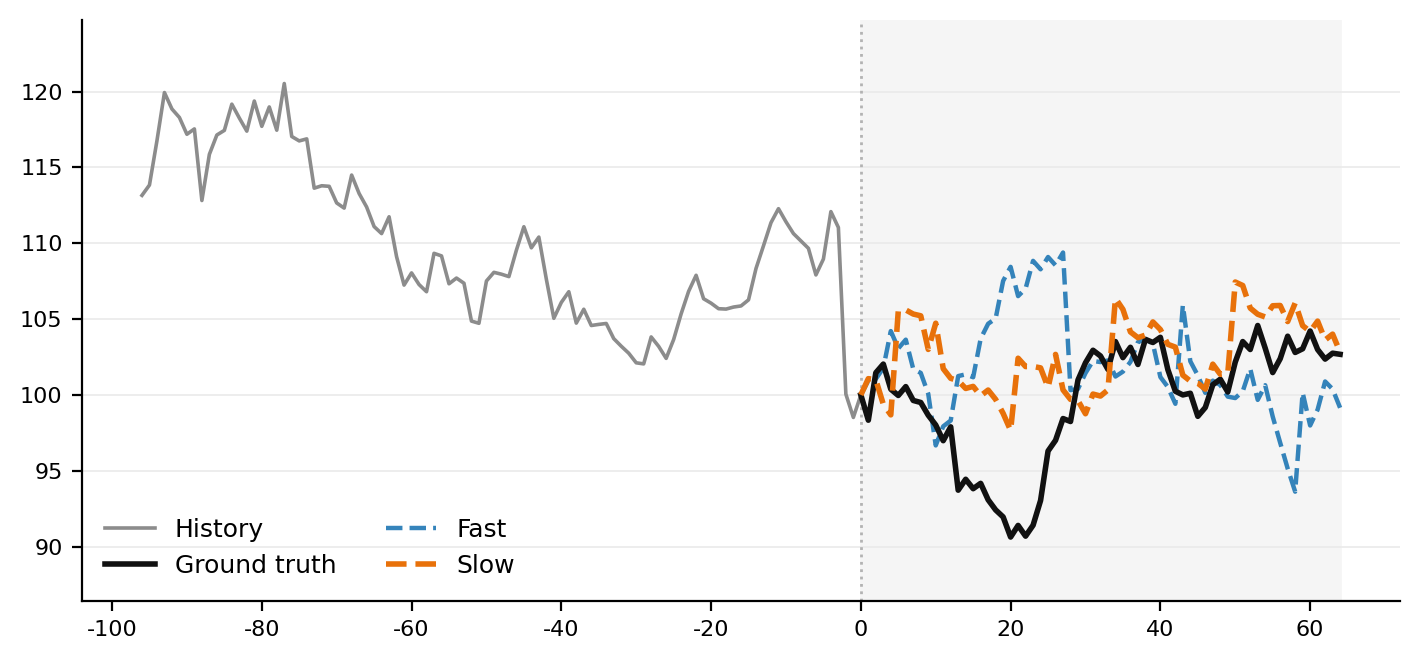}\\[-2pt]
  {\small (b) BMRN, $H{=}64$}
\end{minipage}
\caption{Additional SP500 revision examples. The slow path (orange) conditions
on the fast forecast (blue) and causally available text. In both examples it
removes a substantial portion of the fast path's scale or local-trajectory error,
although it does not reproduce every short-term fluctuation in the realized path
(black). These examples complement the paired aggregate news ablation in
Table~\ref{tab:newsabl}.}
\label{fig:news-examples}
\end{figure*}

\begin{figure*}[!t]
\centering
\begin{minipage}{0.48\textwidth}\centering
  \includegraphics[width=\linewidth]{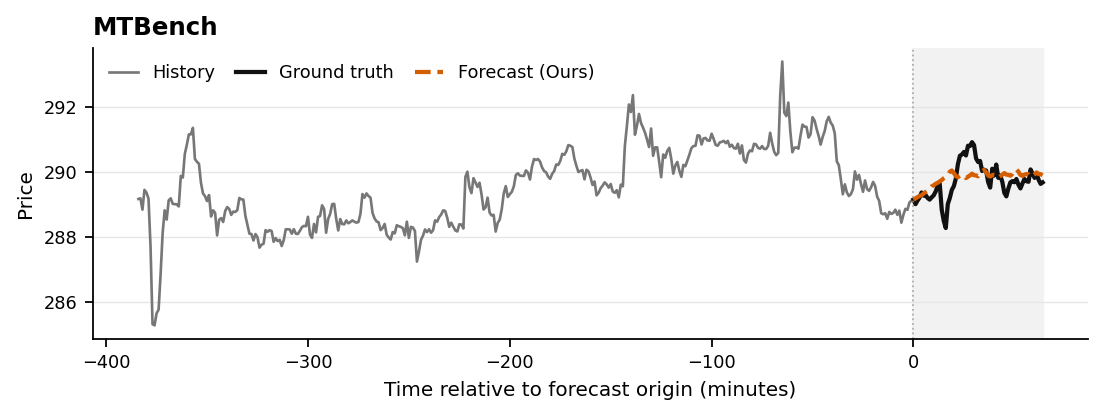}\\[-2pt]
  {\small (a) MTBench (5-minute)}
\end{minipage}\hfill
\begin{minipage}{0.48\textwidth}\centering
  \includegraphics[width=\linewidth]{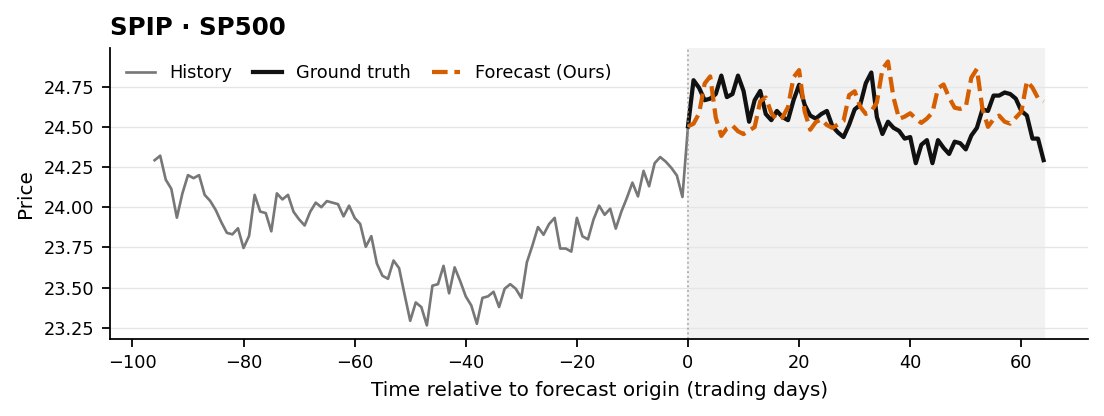}\\[-2pt]
  {\small (b) \textsc{SP500} (daily)}
\end{minipage}

\vspace{0.6em}
\begin{minipage}{0.58\textwidth}\centering
  \includegraphics[width=\linewidth]{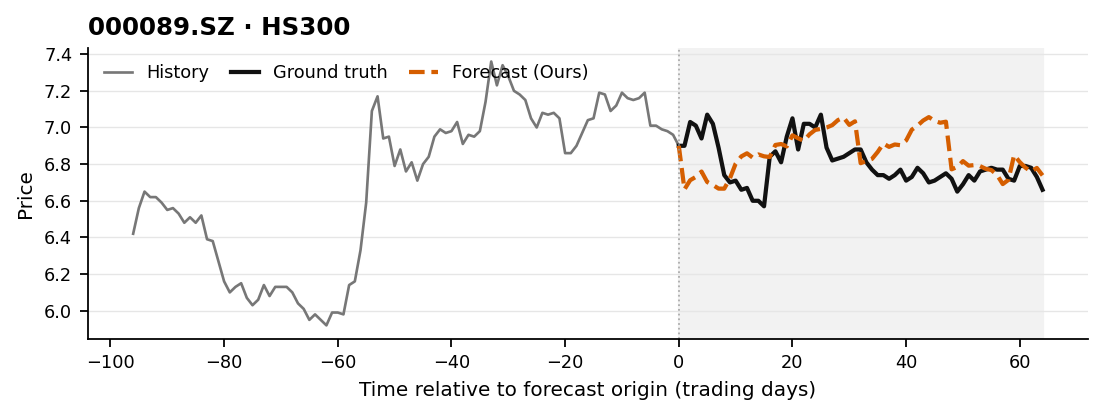}\\[-2pt]
  {\small (c) \textsc{HS300} (daily)}
\end{minipage}

\caption{Representative slow-path forecasts (orange dashed) against the realized
future (black) following the observed history (gray); the dotted vertical line
marks the forecast origin. The three examples cover five-minute MTBench and daily
SP500 and HS300 data.}
\label{fig:forecast}
\end{figure*}

\subsection{Tokenizer Analysis}
\label{sec:afe}

Two design choices underlie the scale--shape tokenizer. First, each RVQ codeword
maps to one embedding row in the extended vocabulary, so highly imbalanced
assignments produce correspondingly imbalanced training of the added vocabulary.
Our Adaptive Frequency‑Equalizing RVQ (AFE‑RVQ) periodically splits overloaded
latent clusters and relocates their sub‑centers into starved slots.

Under the controlled $3\times1024$ setting, AFE‑RVQ preserves reconstruction performance
($0.0783\!\to\!0.0781$), while achieving a modest improvement in finite‑budget active‑code fraction
($0.839\!\to\!0.854$).

The primary benefit of AFE‑RVQ lies in mitigating skewed codeword usage.
Peak load — the maximum assignment count relative to uniform expectation — drops drastically across all residual layers.
For $L_0$, the worst‑case codeword load falls from $7.9\times$ to $3.9\times$; for $L_1$ from $5.0\times$ to $2.6\times$; for $L_2$ from $3.4\times$ to $2.1\times$.
Extreme peak load means a small subset of codewords dominate training updates, harming embedding learning for under‑utilized entries.

\subsection{Qualitative Analysis}

Figure~\ref{fig:forecast} shows stable slow-path forecasts across five-minute
MTBench and daily U.S.\ and Chinese equities. Figure~\ref{fig:modeexamples}
separately visualizes the progression from Fast to SFT and RL. Figure~\ref{fig:news-examples} adds two SP500
windows in which slow revision reduces a substantial fast-path error.
Figure~\ref{fig:cotcards} presents two directional correction cases with
accompanying news and model-generated chain-of-thought rationales.
These examples illustrate individual revisions; aggregate effects on forecast accuracy and direction are reported in Tables \ref{tab:ablation} and \ref{tab:newsabl}.

\begin{figure*}[!t]
\centering
\begin{minipage}{0.48\textwidth}
\setlength{\fboxsep}{6pt}
\setlength{\fboxrule}{0.7pt}
\fcolorbox{green!45!black}{green!3}{%
  \parbox[t][0.20\textheight][t]{\dimexpr\linewidth-2\fboxsep-2\fboxrule\relax}{%
    \footnotesize
    {\color{green!38!black}\bfseries COPX (Copper Miners ETF)\, \textbullet\ U.S.\ market\, \textbullet\ bullish ($H{=}64$)}\par\smallskip
    \textcolor{red!65!black}{\textbf{Fast $-3.3\%$}}~$\longrightarrow$~\textcolor{green!50!black}{\textbf{Slow $+5.9\%$}}~$\longrightarrow$~\textbf{Real $+5.2\%$}\par\smallskip
    \textbf{News.} An approximately \$131.4M weekly inflow into the Copper Miners ETF
    (COPX)---a $4.9\%$ increase in outstanding units---while several underlying miners
    are slightly lower on the day.\par\smallskip
    \textbf{Reasoning}.
    (1)~The large ETF inflow signals rising institutional interest in copper miners, despite mixed
    short-term moves in individual constituents.
    (2)~Net inflows indicate positive sentiment; if demand fundamentals hold, gradual upward pressure should
    continue.
    (3)~I expect COPX to rise moderately over the next 64 trading
    days, roughly $+5\%$ to $+10\%$.\par\smallskip
    \textbf{Outcome.} \textsc{Fast} wrongly predicts a slight decline; the
    news-grounded revision flips the forecast upward and nearly matches the realized
    $+5.2\%$.
  }
}
\end{minipage}\hfill
\begin{minipage}{0.48\textwidth}
\setlength{\fboxsep}{6pt}
\setlength{\fboxrule}{0.7pt}
\fcolorbox{red!55!black}{red!3}{%
  \parbox[t][0.20\textheight][t]{\dimexpr\linewidth-2\fboxsep-2\fboxrule\relax}{%
    \footnotesize
    {\color{red!55!black}\bfseries Kuang-Chi Technologies (002625.SZ)\, \textbullet\ Chinese market\, \textbullet\ bearish ($H{=}64$)}\par\smallskip
    \textcolor{green!50!black}{\textbf{Fast $+27.6\%$}}~$\longrightarrow$~\textcolor{red!65!black}{\textbf{Slow $-3.0\%$}}~$\longrightarrow$~\textbf{Real $-6.4\%$}\par\smallskip
    \textbf{News.} The controlling shareholder and its concert parties announce a plan
    to reduce their stake by up to $2\%$ of shares outstanding.\par\smallskip
    \textbf{Reasoning}.
    (1)~A controlling-shareholder sell-down raises concerns about valuation
    and outlook; with the price already in decline, it can intensify selling pressure.
    (2)~Share reductions are generally read as a negative signal, driving
    near-term outflows and weaker sentiment.
    (3)~Over the next 64 trading days
    I expect continued downward pressure, about $-5\%$ to $-7\%$.\par\smallskip
    \textbf{Outcome.} \textsc{Fast} confidently predicts a large rise; reading the
    insider sell-down as bearish, the revision corrects the sign and tracks the
    realized $-6.4\%$.
  }
}
\end{minipage}

\vspace{0.4em}
\caption{Interpretable revisions: two representative cases where the \textsc{Slow} reviser reads the news, reasons explicitly, and corrects the direction (and sign) of the \textsc{Fast} forecast while approximately matching the realized magnitude. Such directional corrections are illustrative rather than universal; consistent with the news ablation (Table~\ref{tab:newsabl}), text helps where it carries information beyond the price path.}
\label{fig:cotcards}
\end{figure*}

\section{Limitations}
\label{sec:limitations}
This work has two main limitations. First, the beneficial effect of textual news is heterogeneous across different market domains and forecasting settings. While news information effectively improves forecasting performance in most scenarios, the magnitude of its contribution varies across asset types and prediction horizons, leading to marginal or mixed improvements under certain conditions.

Second, the high-performance slow deliberation mode suffers from limited inference speed and increased computational overhead. Compared with the efficient fast forecasting path, the slow path requires additional reasoning token generation, which substantially increases inference latency. This limits the deployment efficiency of the slow mode in time-sensitive real-world scenarios. Optimizing the reasoning efficiency of the deliberative forecasting pipeline remains a promising direction for future work.

\section{Conclusion}
\label{sec:conclusion}
We introduced \textsc{DualCast}, a dual-path language-model framework for financial time-series forecasting. The framework represents each return patch via disentangled summary and residual shape tokens, and leverages AFE-RVQ to balance codeword usage while maintaining high reconstruction fidelity. Based on a frozen Qwen3-8B backbone, DualCast supports two complementary inference modes: a fast financial token decoder for efficient numerical forecasting, and a LoRA-enhanced slow revision module that incorporates causal news information and refines fast predictions. The slow path is initialized via supervised fine-tuning and further optimized with return-space GRPO to improve revision quality.
Extensive experiments across minute-level U.S. equities, daily U.S. and Chinese equities, and weekly energy markets demonstrate that the slow revision mode achieves superior performance, especially over long forecasting horizons. Ablation studies validate the effectiveness of news-guided revision and reinforcement learning optimization. News inputs consistently reduce point prediction errors across most dataset-horizon settings and deliver prominent directional improvements on energy assets.

Empirical results verify our core design paradigm: multimodal financial forecasting can be effectively formulated as an optional news-based revision upon strong numerical base predictions. Future work will explore uncertainty-aware evaluation, generalized market adaptation, and learnable routing strategies to further balance prediction accuracy and inference efficiency.

\section{Acknowledgement}
This work was supported by the Science and Technology Program of Qingdao Municipality under Grant 25-1-1-gjgg-32-gx.
\FloatBarrier





\bibliographystyle{elsarticle-num}
\bibliography{refs}

\end{document}